\documentclass[conference]{IEEEtran}

\usepackage{booktabs}
\usepackage{wrapfig}
\usepackage{graphicx}
\graphicspath{{./}}
\usepackage[table]{xcolor}
\usepackage{tabularx}
\usepackage{array}
\usepackage{booktabs}
\usepackage{amsmath}
\usepackage{amssymb}
\usepackage{multirow}
\usepackage{makecell}
\usepackage{subcaption}
\usepackage{xcolor}
\usepackage{tikz}
\usetikzlibrary{positioning,fit,arrows.meta,calc}

\usepackage{caption}
\title{STeP: Signal Temporal Logic for Precise Specifications for Action Generation with Vision Language Models}

\author{Kasra Torshizi$^{*}$, Anukriti Singh$^{*}$, Sidharth Mathur, Khuzema Habib, Leo Du, Pratap Tokekar\\}

\begin{document}
\maketitle
\let\thefootnote\relax\footnotetext{The authors are with the Department of Computer Science, University of Maryland, College Park. \\ $^{*}$Equal contribution.}

\begin{abstract}

Natural-language robot instructions often specify more than a coarse task goal:
they may impose spatial, temporal, and logical requirements that must remain
satisfied throughout execution. We present \textsc{STeP}, a
specification-based agentic framework that uses Signal Temporal Logic (STL) as
an explicit interface between high-level language reasoning and low-level robot
execution. Rather than encoding such requirements implicitly in a learned
policy, \textsc{STeP} formalizes them as task specifications that can be
decomposed across multi-stage manipulation, enforced during execution,
monitored online, and used as structured feedback for replanning. We evaluate
\textsc{STeP} on standard LIBERO and LIBERO-PRO, and introduce
\textsc{LIBERO-Constrained}, a new benchmark for manipulation tasks with
spatial, temporal, and logical requirements, together with real-world tabletop
experiments. On LIBERO-PRO, \textsc{STeP} retains $54\%$--$89\%$ success across
five of six evaluated perturbation settings, substantially outperforming VLA
and code-as-policy baselines under distribution shift. On
\textsc{LIBERO-Constrained}, \textsc{STeP} achieves $80\%$ safe success across
49 task--constraint instances; on real-world tasks, it improves safe success
over a specification-free model-based baseline across all four task
categories, with gains of up to $45$ percentage points. These results support
explicit formal specifications as a practical interface between
foundation-model reasoning and reliable robot execution.
\end{abstract}


\section{Introduction}
\label{sec:introduction}

Robot policies conditioned on natural language must do more than satisfy the
broad intent of an instruction. Many natural-language instructions contain
spatial, temporal, and logical requirements that determine whether the
resulting behavior is actually correct: an object may need to be placed at a
specified metric offset, a maneuver may need to occur within a timing window,
or an action may be conditioned on an event elsewhere in the task. Moreover,
without an explicit
measure of requirement satisfaction during execution, it is difficult to
determine which requirement is being approached or violated, provide precise
feedback about a failure, or use that feedback to improve subsequent
execution. In this paper, we present a specification-based agentic framework
for language-conditioned manipulation that makes task requirements explicit
during execution and uses their execution status to provide structured
feedback for replanning.

\begin{figure}[h]
    \centering
    \includegraphics[width=1.0\linewidth]{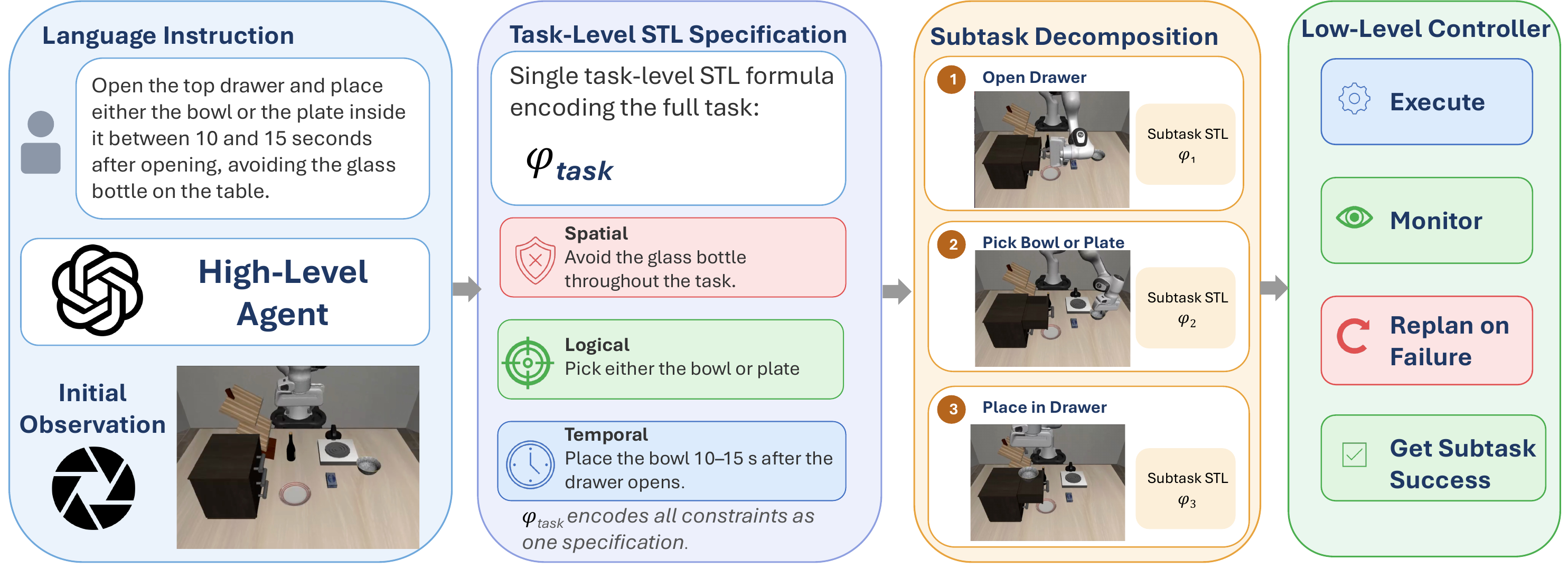}
    \caption{Overview of \textsc{STeP}. System 2 converts a natural-language
instruction into an STL task specification, which is decomposed into executable
subtasks while preserving task constraints. STL robustness supports monitoring
and replanning during execution. Figure~\ref{fig:full_pipeline} shows the full pipeline for a similar task.}
    \label{fig:motivating}
\end{figure}

Two dominant paradigms for language-conditioned manipulation address parts of
this problem but represent task requirements differently.
Vision-Language-Action (VLA) models map instructions and observations directly
to actions through an end-to-end policy
~\cite{kim2024openvlaopensourcevisionlanguageactionmodel,
black2026pi0visionlanguageactionflowmodel,black2025pi05visionlanguageaction,
nvidia2025gr00tn1openfoundation}. While these models generalize across diverse
manipulation behaviors, spatial, temporal, and logical requirements are
typically represented implicitly within the learned policy. Code-as-Policies
and related methods instead generate executable code or sequences of skills
over a robot API
~\cite{liang2023codepolicieslanguagemodel,
singh2022progpromptgeneratingsituatedrobot,ahn2022icanisay}. These approaches
provide explicit procedural structure, but task requirements and the procedure
used to satisfy them remain coupled within the generated program. We are
instead interested in separating \emph{what conditions the robot should
satisfy} from \emph{how those conditions are achieved}, so that the
requirements themselves can be preserved and evaluated throughout execution.
Temporal logic~\cite{4567924} provides a natural representation for this
purpose. Signal Temporal Logic (STL)
~\cite{10.1007/978-3-540-30206-3_12}, in particular, can express spatial,
temporal, and logical conditions over continuous-valued signals. Beyond
Boolean satisfaction, STL provides a quantitative robustness value measuring
the margin by which a requirement is satisfied or violated
~\cite{donze2010robust,sadraddini2015robust}.
Recent work connects temporal logic with language-conditioned robot behavior,
translating instructions into Linear Temporal Logic (LTL) or STL
specifications for conditioning
policies, planning, or verification~\cite{wang2026logicvla,
he2022deepstlenglishrequirements,5e25b65f791049f09ef72f9930c86672,
fang2025enhancingtransformationnaturallanguage,
choi2026reachabilitybasedtemporallogicverification,chen2024autotamp,
vandelaar2024vernacopter}. Multi-stage manipulation, however, requires more
than translation. A high-level agent must reason over requirements spanning
several stages while the controller executes only the current subtask,
persistent constraints must stay active across those stages, and execution
feedback must remain tied to the requirement it came from.

We address this challenge with \textsc{STeP}, a specification-based agentic
framework that connects a high-level reasoning agent (System 2) with a low-level
model-based controller (System 1) through an STL task specification. System 2 reasons deliberatively
about the task and produces the specification, while System 1 executes against
that specification in closed loop. System 2 interprets the
language instruction, constructs the task-level specification, and decomposes
it into an ordered sequence of subtask specifications while preserving
requirements, such as safety constraints, that remain active across multiple
stages. This decomposition follows the spirit of LTL/STL formula progression
~\cite{bacchus2000using} and related decomposition methods
~\cite{kapoor2024safe,yu2022mpc}. Figure~\ref{fig:motivating} illustrates the
overall flow.


System 1 executes each subtask using a primarily model-based
approach, while incorporating learned components where useful. It operates on
an explicit task-relevant scene state, including the robot configuration,
object identities and poses, and geometric relations needed by the active
predicates, provided through the system's perception and state-estimation
interface. For example, grasp selection can be handled by a learned perception
module, while motion is generated and controlled using model-based planning
and control. The active STL specification is incorporated into this execution
process so that the generated behavior accounts explicitly for the requirements
identified by System 2
~\cite{raman2014model,sadraddini2015robust}. At the same time, an online STL
monitor evaluates robustness during execution. When a requirement is violated,
the corresponding requirement, violation time, and robustness margin are
returned to System 2, providing structured context for replanning
without discarding the original task specification.

To evaluate these capabilities systematically, we introduce
LIBERO-Constrained, a benchmark built on LIBERO that augments
manipulation tasks with spatial, temporal, and logical requirements, together
with controlled failures for studying replanning. We complement this benchmark
with standard LIBERO, LIBERO-PRO, and real-world tabletop experiments. Our
evaluation addresses three questions:

\noindent\textbf{RQ1: General Use as a Multi-Function Model.}
Can \textsc{STeP} solve diverse manipulation tasks competitively with
code-as-policy agents and VLAs?

\noindent\textbf{RQ2: Precise Language Following.}
Does STL-guided execution improve satisfaction of spatial, temporal, and
logical requirements without sacrificing task success?

\noindent\textbf{RQ3: Few-Shot Replanning.}
Does preserving the specification and robustness feedback help
\textsc{STeP} recover from execution failures while staying grounded in the
instruction?

Across these evaluations, \textsc{STeP} remains competitive on general
manipulation and is substantially more robust under distribution shift: on
LIBERO-PRO it retains $54\%$--$89\%$ success in five of six settings, while no
baseline exceeds $38\%$. On \textsc{LIBERO-Constrained}, \textsc{STeP}
achieves $80\%$ safe success across 49 task--constraint instances, compared
with $14\%$ for SmolVLA, while reducing the violation rate from $66\%$ to
$20\%$.

\section{Related Work}
\label{sec:related_work}

Our work lies at the intersection of language-based robot planning,
temporal-logic control, and closed-loop agentic reasoning. We position
\textsc{STeP} as a modular alternative to approaches that place increasingly
more of the perception-to-action pipeline inside a single learned policy.

\paragraph*{Language-Based Planning}
Language models have been used for high-level skill selection, executable
policy generation, and geometric task reasoning. SayCan~\cite{ahn2022icanisay}
grounds language-model predictions in robot affordances, while
Code-as-Policies~\cite{liang2023codepolicieslanguagemodel} generates executable
programs over robot APIs. Other approaches connect language reasoning to
geometric optimization: VoxPoser~\cite{huang2023voxposercomposable3dvalue}
uses language-derived 3D value maps, while ReKep
~\cite{huang2024rekepspatiotemporalreasoningrelational} optimizes relational
keypoint constraints. \textsc{STeP} follows this modular view, but introduces
an explicit temporal-logic specification that persists between high-level
reasoning and low-level execution.

\paragraph*{Temporal Logic for Planning and Learning}
Temporal logic has long been used to express sequencing, safety, reachability,
and invariance requirements in robot planning
~\cite{10.1109/TRO.2009.2030225,raman2014model,sadraddini2015robust}.
Recent work also connects natural language with formal specifications:
DeepSTL~\cite{he2022deepstlenglishrequirements} translates language to STL,
while AutoTAMP~\cite{chen2024autotamp} and
VernaCopter~\cite{vandelaar2024vernacopter} use temporal logic as an
intermediate representation for planning and control. In parallel, formal
specifications have been incorporated directly into learned controllers and
policies~\cite{meng2023signal,pmlr-v235-guo24j}. \textsc{STeP} instead uses STL
as a shared interface across reasoning, decomposition, model-based execution,
monitoring, and replanning.

\paragraph*{Temporal-Logic-Conditioned Robot Policies}
The closest recent approaches integrate temporal-logic requirements directly
into learned action generation. Logic-VLA~\cite{wang2026logicvla} adapts a VLA
using STL-conditioned training, while Vision-TL-Action
~\cite{liu2026visiontlaction} conditions a learned trajectory generator on
visual observations and a temporal-logic representation. In contrast,
\textsc{STeP} keeps the high-level reasoning agent, formal specification, and
low-level controller decoupled. The same STL specification can therefore be
decomposed across task stages, used by model-based planning and control,
monitored during execution, and returned to the reasoning agent as structured
feedback without retraining the low-level policy.

\paragraph*{Closed-Loop Reasoning and Failure Recovery}
A complementary line of work uses foundation models for iterative planning and recovery. Inner Monologue~\cite{huang2022innermonologueembodiedreasoning} feeds execution feedback back to a Large Language Model (LLM) planner, REFLECT~\cite{liu2023reflectsummarizingrobotexperiences} uses multimodal experience
to diagnose failures and produce corrective plans, and more recent systems use Vision-Language Model (VLM) based failure detection and test-time reflection for recovery
~\cite{pacaud2025guardian,hong2026reflective}. \textsc{STeP} shares this
closed-loop structure, but grounds feedback in the original formal
specification through requirement-specific STL robustness.
\vspace{-0.5em}
\section{Background}

This section introduces the STL syntax, signal predicates, and quantitative
robustness semantics used by \textsc{STeP}. Consider an environment with state
space $\mathcal{S}$, observation space $\mathcal{O}$, and action space
$\mathcal{A}$. At timestep $t$, the
environment has state $s_t\in\mathcal{S}$, the robot observes
$o_t\in\mathcal{O}$ and executes $a_t\in\mathcal{A}$. The robot is given a
natural-language instruction $\ell\in\mathcal{L}$ and has access to a set of
skills $\{m_1,m_2,\ldots\}$.

\subsection{Signal Temporal Logic}
\label{sec:formulation}

STL provides a formal language for specifying spatial and temporal properties
of robot trajectories. We define STL formulas recursively in Backus--Naur
form~\cite{inbook,10.1007/978-3-540-30206-3_12}:
\begin{align*}
\varphi ::= {} & \top \mid \mu_{j,c} \mid \neg \varphi
\mid \varphi_1 \wedge \varphi_2 \mid \varphi_1 \vee \varphi_2 \\
\mid {} & \varphi_1 \Rightarrow \varphi_2
\mid \mathbf{F}_{[a,b]}\varphi
\mid \mathbf{G}_{[a,b]}\varphi
\mid \varphi_1\,\mathbf{U}_{[a,b]}\,\varphi_2,
\end{align*}
where $\mathbf{F}_{[a,b]}$, $\mathbf{G}_{[a,b]}$, and
$\mathbf{U}_{[a,b]}$ denote ``eventually,'' ``always,'' and ``until'' over
$[a,b]$, respectively. We include implication explicitly,
$\varphi_1\Rightarrow\varphi_2\equiv\neg\varphi_1\vee\varphi_2$, because
conditional structure appears naturally in language instructions. We denote
the set of formulas generated by this grammar by $\Phi$.

\subsection{Signal Predicates}
\label{sec:signal_predicates}

STL formulas are evaluated over task-relevant signals
$z_t=\psi(s_t,o_t)\in\mathcal{Z}$. We assume a predefined predicate library
$\mathcal{P}=\{\mu_1,\ldots,\mu_M\}$, with
$\mu_j:\mathcal{Z}\times\Theta_j\rightarrow\mathbb{R}$.

Given predicate parameters $\theta\in\Theta_j$ specifying the relevant objects,
regions, or other task quantities (e.g., the bowl in \texttt{Grasped}), and a
threshold $c$, an atomic predicate is
$\mu_{j,c}(z_t;\theta):=\mu_j(z_t;\theta)\geq c$, with robustness
\[
\rho(\mathbf{z},t,\mu_{j,c})=\mu_j(z_t;\theta)-c.
\]
The predicate is satisfied when this value is nonnegative. The agent selects
predicates from $\mathcal{P}$ rather than generating arbitrary signal
functions; a generic \texttt{GoalPosition} predicate handles goals not covered
by the predefined set. Predicates used during Model-Predictive Control (MPC)
optimization must be
differentiable or admit a smooth proxy, whereas predicates used only for
symbolic decomposition or discrete monitoring need not be (e.g.,
\texttt{Released}).

\subsection{Robustness}
\label{sec:robustness}

In addition to Boolean satisfaction, STL admits quantitative robustness
semantics. Let $\mathbf{z}=(z_0,z_1,\ldots)$ denote the signal trajectory. The
robustness of formula $\varphi$ at time $t$ is
$\rho(\mathbf{z},t,\varphi)\in\mathbb{R}$: positive values indicate
satisfaction, negative values indicate violation, and the magnitude measures
the margin from the satisfaction boundary. The recursive definitions are
given in Table~\ref{tab:stl_robustness}.

\begin{table}[h]
\centering
\begin{tabular}{ll}
\toprule
Formula $\varphi$ & Robustness $\rho(\mathbf{z}, t, \varphi)$ \\
\midrule
$\mu_{j,c}$ & $\mu_j(z_t) - c$ \\
$\neg \varphi$ & $-\rho(\mathbf{z}, t, \varphi)$ \\
$\varphi_1 \wedge \varphi_2$ & $\min\big(\rho(\mathbf{z},t,\varphi_1),\, \rho(\mathbf{z},t,\varphi_2)\big)$ \\
$\varphi_1 \vee \varphi_2$ & $\max\big(\rho(\mathbf{z},t,\varphi_1),\, \rho(\mathbf{z},t,\varphi_2)\big)$ \\
$\varphi_1 \Rightarrow \varphi_2$ & $\max\big(-\rho(\mathbf{z},t,\varphi_1),\, \rho(\mathbf{z},t,\varphi_2)\big)$ \\
$\mathbf{F}_{[a,b]}\varphi$ & $\displaystyle\max_{t' \in [t+a,\,t+b]} \rho(\mathbf{z}, t', \varphi)$ \\
$\mathbf{G}_{[a,b]}\varphi$ & $\displaystyle\min_{t' \in [t+a,\,t+b]} \rho(\mathbf{z}, t', \varphi)$ \\
$\varphi_1\,\mathbf{U}_{[a,b]}\,\varphi_2$ & $\displaystyle\max_{t' \in [t+a,\,t+b]} \min\Big(\rho(\mathbf{z}, t', \varphi_2),\ \min_{t'' \in [t,\,t')} \rho(\mathbf{z}, t'', \varphi_1)\Big)$ \\
\bottomrule
\end{tabular}
\caption{Recursive definition of STL quantitative robustness semantics.}
\label{tab:stl_robustness}
\end{table}



\section{Method}
\label{sec:method}
\begin{figure*}[t]
    \centering
    \includegraphics[width=\textwidth]{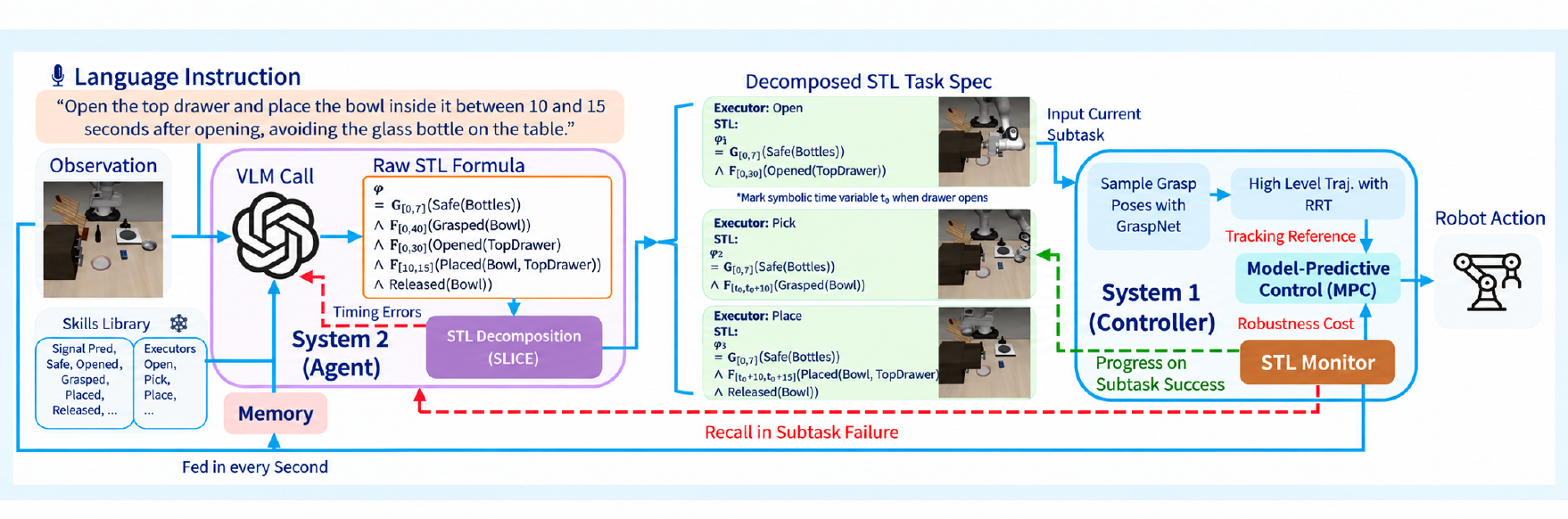}
    \caption{Overview of \textsc{STeP}'s full pipeline on the running drawer
    example. System~2 formalizes the language instruction, current observation,
    and Skills Library into a single STL specification $\varphi$, then decomposes
    it following SLICE~\cite{kapoor2024safe} into ordered subtask specifications
    $\varphi_1$ (\texttt{open}), $\varphi_2$ (\texttt{pick}), and $\varphi_3$
    (\texttt{place}), each paired with its executor. Note that an outer temporal operator's window bounds when its own enclosed goal is witnessed, not any goal nested further inside it, so a nested deadline can extend past the outer window. System~1 samples grasp poses with Contact-GraspNet
    ~\cite{sundermeyer2021contactgraspnet}, plans a reference trajectory with RRT,
    and tracks it using STL-robustness-guided MPC. The STL monitor advances to the
    next subtask on success and triggers recall to System~2 with Memory context on
    failure. MPC runs at 10\,Hz and Memory is updated once per second.}
    \label{fig:full_pipeline}
\end{figure*}

Our framework separates language-conditioned robot planning into two interacting modules. A high-level \textit{System 2} (Agent) formalizes and decomposes the instruction into a sequence of STL-specified subtasks, while a low-level \textit{System 1} (Controller) executes each subtask through classical motion planning. STL serves as the interface between the two systems: System 2 generates STL specifications from language, and System 1 uses these specifications for execution, monitoring, and replanning. Figure~\ref{fig:full_pipeline} showcases a high-level overview. 

\subsection{System 2: Language-to-STL Task Planning}

Given a language instruction $\ell$, the current observation $o$, and a recall context $\mathcal{H}_t$ of prior execution, System 2 queries a VLM formalizer $\Gamma$ to produce a single STL specification for the entire task,
\[
\varphi = \Gamma(\ell, o, \mathcal{H}_t) \in \Phi,
\]
where $\Phi$ is the set of formulas generated by the grammar of Section~\ref{sec:formulation} over a fixed \textit{Skills Library}: a predicate library $\mathcal{P}$ (Section~\ref{sec:signal_predicates}) together with a set of executable \textit{skills} (e.g., \texttt{pick}, \texttt{place}), each realized by a corresponding executor in System 1. Unlike a per-subtask specification, $\Gamma$ only formalizes $\ell$ into one formula over the full task horizon; where subtask boundaries fall is determined afterward from $\varphi$'s own temporal structure, not by the VLM (Figure~\ref{fig:full_pipeline}).


\subsubsection{STL Task Spec}
System 2 is prompted to generate $\varphi$ so that it separates into persistent safety and invariance requirements together with an ordered sequence of achievement goals,
\[
\varphi = \Big(\bigwedge_{i=1}^p \mathbf{G}_{[0,T]}\chi_i\Big) \wedge \big(\varphi_1^{\mathrm{ach}} \wedge \cdots \wedge \varphi_n^{\mathrm{ach}}\big),
\]
where each $\chi_i$ must hold throughout the task (e.g., obstacle avoidance) and each $\varphi_j^{\mathrm{ach}}$ is an $\mathbf{F}_{[a_j,b_j]}$- or $\mathbf{U}_{[a_j,b_j]}$-guarded subgoal, with $[a_j,b_j]$ its own outer window; $\varphi_j^{\mathrm{ach}}$ may itself be a conjunction of several timed checkpoints toward the same goal (e.g., a pre-grasp pose before the grasp itself), which can freely share or overlap in time since they belong to one subtask. Achievement goals are ordered by when they are intended to be satisfied, with consecutive goals' outer windows non-overlapping and non-decreasing ($b_j \le a_{j+1}$) per SLICE's~\cite{kapoor2024safe} timeline-slicing procedure, so that each maps to its own subtask; a goal whose timing depends on an earlier one is instead nested inside it and anchored to that goal's own symbolic time variable, exempting it from this ordering.

Similar to~\cite{bacchus2000using,kapoor2024safe,yu2022mpc}, we define a decomposition operator
\[
D(\varphi) = (\varphi_1,\ldots,\varphi_n), \quad \varphi_j := \varphi_j^{\mathrm{ach}} \wedge \bigwedge_{i=1}^p \mathbf{G}_{[0,T]}\chi_i,
\]
so every subtask formula carries its own achievement goal together with the full set of persistent constraints, rather than leaving safety and invariance to any single subtask. System 1 advances from $\varphi_j$ to $\varphi_{j+1}$ once $\rho(\mathbf{z},t,\varphi_j^{\mathrm{ach}}) \geq 0$ (Table~\ref{tab:stl_robustness}), the same progression rule by which a satisfied subformula is dropped in favor of the residual formula over the remaining future~\cite{bacchus2000using}. Figure~\ref{fig:full_pipeline} shows $\varphi$ instantiated for our running drawer example, together with the result of applying $D$ to it.

Each subtask's achievement goal names one predicate from $\mathcal{P}$ (e.g., \texttt{Grasped}, \texttt{OverGoal}, \texttt{Placed}). A fixed predicate-to-skill map $\kappa$, defined once alongside the Skills Library and extended whenever a new skill is added, grounds this predicate to the executor that realizes it (e.g., $\kappa(\texttt{Grasped})=\texttt{pick}$), so skill selection follows from which predicate a subtask names rather than requiring the agent to specify it directly. A natural alternative, closer to Code-as-Policies~\cite{liang2023codepolicieslanguagemodel}, would have System 2 synthesize each predicate's evaluation code directly rather than select from a fixed table; we use a fixed $\kappa$ instead so that every predicate stays checkable against the grammar of Section~\ref{sec:formulation}, at the cost of restricting skills to a closed library. The result, $\big((m_1,\varphi_1),\ldots,(m_n,\varphi_n)\big)$ with $m_j = \kappa(\varphi_j^{\mathrm{ach}})$, is the task specification handed to System 1.

\subsection{System 1: STL-Guided Execution and Monitoring}

System 1 executes the active subtask formula $\varphi_j$ from System 2 and returns the robot action for the current timestep.

\subsubsection{STL-Guided MPC}
For each subtask, a sampling-based Rapidly-exploring Random Tree (RRT) planner computes a collision-free reference path $q^{\mathrm{ref}}=(q_0^{\mathrm{ref}},\ldots,q_K^{\mathrm{ref}})$ to the subtask's target. For pick subtasks, this target is a grasp pose from Contact-GraspNet~\cite{sundermeyer2021contactgraspnet}, a learned grasp-pose network. Once an object is grasped, it is treated as part of the robot's rigid body until it is released: its collision geometry is approximated by a small set of spheres fit to its extent at grasp time, which then move rigidly with the end-effector and are checked against the rest of the scene during subsequent planning and tracking, rather than re-evaluating its exact mesh at every query. An MPC controller then tracks $q^{\mathrm{ref}}$ in closed loop while directly optimizing STL robustness, letting the executed trajectory depart from the reference whenever doing so improves constraint satisfaction: this is where STL actually enters low-level control, rather than only being checked after the fact. The controller's own state is restricted to the robot's joint configuration $q_t$, a subset of the full environment state $s_t$; in order to enact MPC, we have to prescribe a dynamics model $
q_{t+1}=f(q_t,a_t)$,
which predicts how the joint configuration evolves under a low-level action $a_t$. Here $f$ is manually prescribed rather than learned: $f$ is a single-integrator rollout, $q_{t+1}=q_t+\eta\,a_t$, where $\eta$ is a fixed per-step scale bounding the maximum joint delta. At each timestep, MPC optimizes an action sequence over a finite horizon $H$, which is typically shorter than a subtask's own time window (Section~\ref{sec:limitations} discusses the resulting timing behavior and limitations with this approach), executes only the first action, and replans at the next timestep. Given the active subtask formula $\varphi_j$ from $D(\varphi)$ and the current reference index $k_0$ along $q^{\mathrm{ref}}$, the controller solves
\begin{align*}
a_{t:t+H-1}^{\star}
={}&
\arg\min_{a_{t:t+H-1}}
\sum_{k=t}^{t+H-1} \big(w_k \big\|q_{k|t}-q^{\mathrm{ref}}_{k_0+k-t}\big\|_2^2\\
&+c_u(a_k)
-
\lambda \rho(\mathbf{z}_{t:t+H|t},t,\varphi_j)\big),
\end{align*}
subject to $
q_{k+1|t}=f(q_{k|t},a_k), z_{k|t}=\psi(q_{k|t},o_t),
$
where $w_k$ is a tracking weight that ramps up over the horizon and is largest at the terminal step $k=t+H-1$, and $c_u$ penalizes control effort and non-smoothness between consecutive actions. Here, $\mathbf{z}_{t:t+H|t}=(z_{t|t},\ldots,z_{t+H|t})$ is the predicted signal trajectory used to evaluate STL robustness. As robustness semantics of Table~\ref{tab:stl_robustness} are built from \textsc{max} and \textsc{min}, which are not differentiable, the controller optimizes a smooth approximation of them,
\[
\begin{aligned}
\max(\mathbf{x}) \approx \frac{\sum_{i=1}^{n} x_i \exp(\beta x_i)}
{\sum_{i=1}^{n} \exp(\beta x_i)}, \qquad \min(\mathbf{x}) = - \max(-\mathbf{x})
\end{aligned}
\]
where $\beta > 0$ controls the sharpness of the approximation (in our implementation, we set $\beta = 30$). Since the dynamics model predicts future states but not future observations, future signals are computed using the predicted states and the current observation $o_t$. 

\subsubsection{Monitoring and Replanning}
During execution, an STL monitor evaluates the robustness of the active subtask formula. This robustness value provides a continuous measure of progress and constraint satisfaction. If robustness falls below a threshold, System 1 triggers a recall to System 2 using the current execution context; in practice, this threshold applies only to persistent safety and failure-type predicates (e.g., collision avoidance, an object slipping from the gripper), since ordinary task predicates dip negative transiently during a normal approach. As an additional trigger, the subtask's own achievement goal is monitored against its formula's time window: System 1 also declares it failed once the window's upper bound elapses without the goal predicate having been satisfied, which likewise triggers a recall. This recall mechanism allows the system to repair or revise the plan while remaining grounded in the original task specification. The same monitor detects completion: once the active formula is satisfied, the system advances to the next subtask.

\subsubsection{Memory}

To provide context for monitoring and replanning, the system maintains a rolling history log during execution. The log is updated at a fixed rate and stores the most recent $L$ records, $\mathcal{H}_t = \{r_{t-L+1},\ldots,r_t\}$, where each record contains the current visual observation $o_t$, robot state (joint positions, gripper positions, impedances), natural-language description of the subtask, stage, and STL robustness values. When System 1 triggers a recall, this history log is passed to System 2 together with the current STL task specification and prior plan. This allows System 2 to reason over recent execution progress and failures without discarding the original task structure.

\section{Evaluation} 
\label{sec:results_discussion}





\begin{figure}[t]
    \centering

    \includegraphics[width=0.87\linewidth]{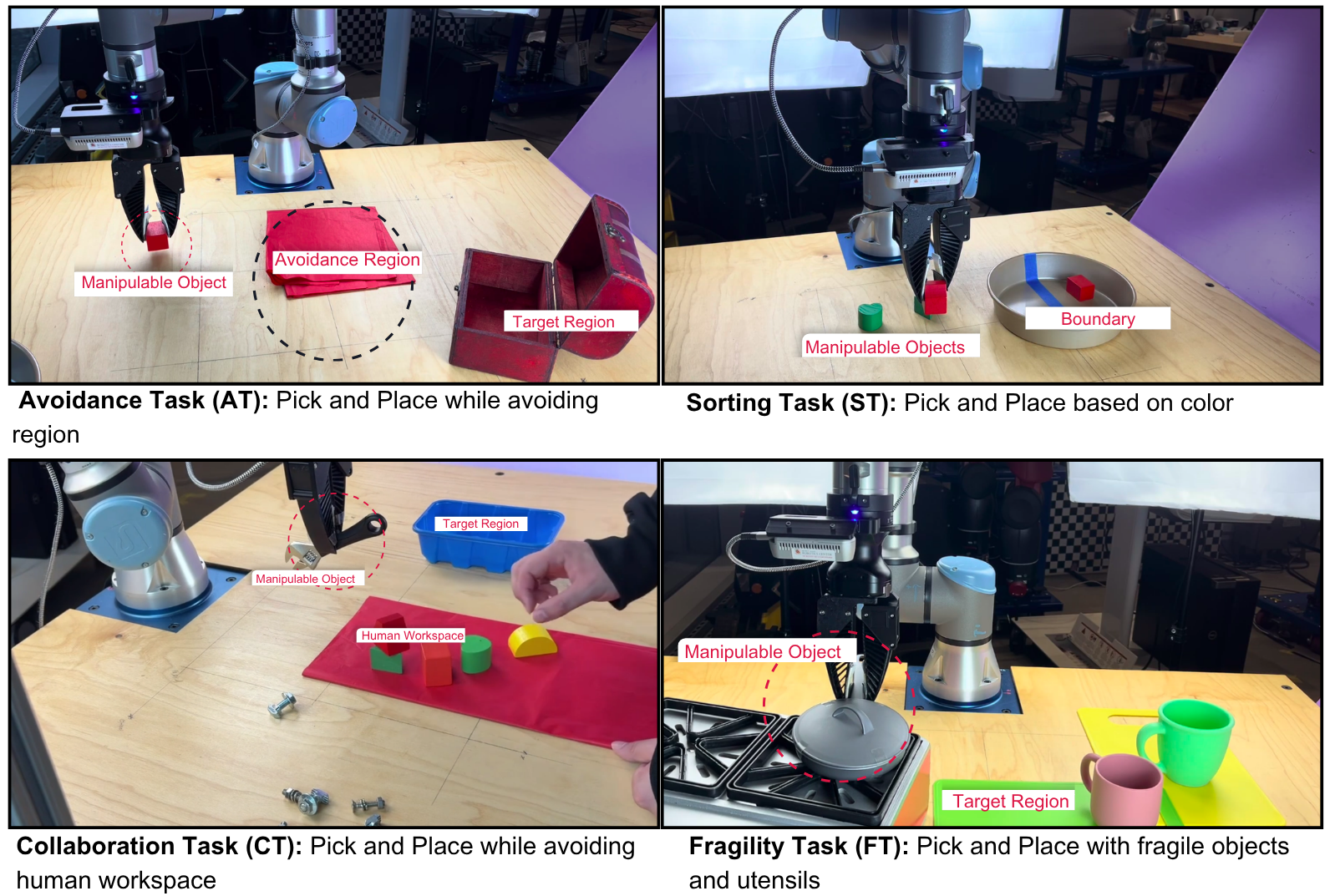}

    \vspace{0.4em}

    \includegraphics[width=0.78\linewidth]{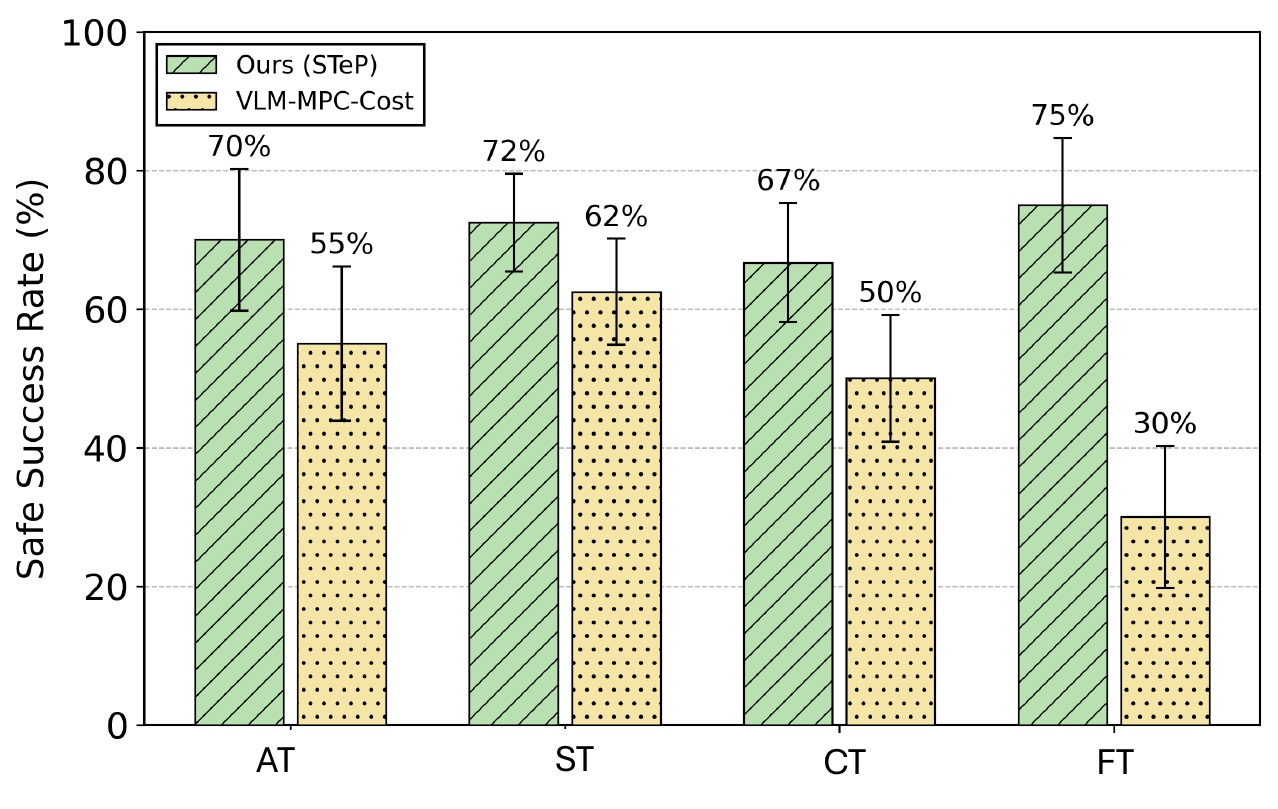}

    \caption{Real-world evaluation on four pick-and-place task categories.
    \textbf{Top:} representative examples of the Avoidance (AT), Sorting (ST),
    Collaboration (CT), and Fragility (FT) tests, illustrating the corresponding
    task-specific constraints. \textbf{Bottom:} safe success rate of
    \textsc{STeP} and \textsc{VLM-MPC-Cost}; \textsc{STeP} improves safe success
    across all four categories, with the largest gain on the fragility task.}
    \label{fig:real_world_combined}
\end{figure}

We evaluate our framework around the three presented research questions described in Section~\ref{sec:introduction}. 




\paragraph*{Experimental Setup}
 On our LIBERO-based benchmarks, we compare \textsc{STeP} against a variety of VLAs; on LIBERO-PRO specifically, we additionally compare against a code-as-policy agent, \textsc{CaP-Agent0}. In the real-world tabletop experiments, we instead compare against an ablated version of our own framework with the STL components removed. In this baseline, which we refer to as \textsc{VLM-MPC-Cost}, System 2 outputs a direct MPC cost function using the available predicate set, together with a failure threshold for recall, rather than compiling the task into STL formulas. The history log records MPC costs instead of STL robustness traces. The VLM calls were always made using GPT-5.4. To resolve objects and their locations in LIBERO we read privileged simulator state, while on the real UR3e we instead obtain it using a perception module that includes Set-of-Mark (SoM)~\cite{yang2023setofmarkpromptingunleashesextraordinary} prompting to segment and label every object in the scene and a RealSense depth camera to recover each object's 3D centroid, calibrated into the robot's coordinate frame. As a note, our real-world implementation uses a fixed grasp pose relative to the object instead of the GraspNet used in the other experiments. In our experiments, we had an MPC Horizon $H=16$ which is a $1.6s$ lookahead.
\paragraph*{\textbf{LIBERO-Constrained Benchmark}}
\label{sec:libero_constrained}

To evaluate whether a robot satisfies requirements beyond nominal task
completion, we introduce \textsc{LIBERO-Constrained}, a set of 49
task--constraint instances built from LIBERO. The benchmark spans four
constraint families: \emph{spatial} constraints such as keep-out regions,
metric offsets, and relative placement; \emph{logical} constraints such as
ordering, preconditions, and leaving specified objects untouched;
\emph{temporal} constraints such as waiting, dwell-time, and stage deadlines;
and \emph{combined} tasks that compose requirements from multiple families.
The benchmark contains 23 spatial, 8 logical, 11 temporal, and 7 combined
instances. Representative examples are shown in Figure~\ref{fig:libero_constrained_results};
the complete task specification is provided on the project webpage.

\subsection{RQ1: General Manipulation Capability}
\label{sec:rq1}

\begin{table}[t]
\centering
\scriptsize
\setlength{\tabcolsep}{4pt}
\begin{tabular}{lccccc}
\toprule
\textbf{Method} & \textbf{Spatial} & \textbf{Object} & \textbf{Goal} & \textbf{Long} & \textbf{Total} \\
\midrule
OpenVLA~\cite{kim2024openvlaopensourcevisionlanguageactionmodel} & 0.85 & 0.88 & 0.79 & 0.54 & 0.77 \\
$\pi_{0.5}$~\cite{black2025pi05visionlanguageaction} & 0.99 & 0.98 & 0.98 & 0.92 & 0.97 \\
SmolVLA~\cite{shukor2025smolvla} & 0.90 & 0.96 & 0.92 & 0.71 & 0.87 \\
\midrule
\textsc{STeP} (Ours) & 0.87 & 0.92 & 0.76 & 0.74 & 0.82 \\
\bottomrule
\end{tabular}
\caption{Task success rate on standard (unperturbed) LIBERO, per suite.
Baseline results are self-reported by OpenVLA, $\pi_{0.5}$, and SmolVLA;
\textsc{STeP} results are from our own rollouts. Standard LIBERO evaluates
nominal task execution, while Table~\ref{tab:libero_pro_results} evaluates
the same setting under position and task perturbations.}
\label{tab:libero_standard_results}
\end{table}

\begin{table}[t]
\centering
\scriptsize
\setlength{\tabcolsep}{3.5pt}
\begin{tabular}{lcccccc}
\toprule
& \multicolumn{2}{c}{\textbf{Spatial}} & \multicolumn{2}{c}{\textbf{Object}} & \multicolumn{2}{c}{\textbf{Goal}} \\
\cmidrule(lr){2-3} \cmidrule(lr){4-5} \cmidrule(lr){6-7}
\textbf{Method} & \textbf{Pos} & \textbf{Task} & \textbf{Pos} & \textbf{Task} & \textbf{Pos} & \textbf{Task} \\
\midrule
OpenVLA~\cite{kim2024openvlaopensourcevisionlanguageactionmodel} & 0.00 & 0.00 & 0.00 & 0.00 & 0.00 & 0.00 \\
$\pi_0$~\cite{black2026pi0visionlanguageactionflowmodel} & 0.00 & 0.00 & 0.00 & 0.00 & 0.00 & 0.00 \\
$\pi_{0.5}$~\cite{black2025pi05visionlanguageaction} & 0.20 & 0.01 & 0.17 & 0.01 & 0.38 & 0.00 \\
\textsc{CaP-Agent0}~\cite{fu2026capx} & 0.12 & 0.14 & 0.22 & 0.18 & 0.26 & 0.17 \\
\midrule
\textsc{STeP} (Ours) & 0.73 & 0.69 & 0.85 & 0.89 & 0.54 & 0.16 \\
\bottomrule
\end{tabular}
\caption{Task success rate on LIBERO-PRO~\cite{zhou2025liberopro} under position (Pos) and task (Task) perturbations. Baseline rows are as reported in CaP-X's Table 2~\cite{fu2026capx}; the \textsc{STeP} row is our own rollouts.}
\label{tab:libero_pro_results}
\end{table}

We first evaluate \textsc{STeP} as a general-purpose multi-function model on standard LIBERO and LIBERO-PRO, which perturbs the same tasks' initial positions and task phrasing.

On standard LIBERO (Table~\ref{tab:libero_standard_results}), \textsc{STeP} reaches a $82\%$ overall success rate, between OpenVLA ($77\%$) and SmolVLA ($87\%$) and below $\pi_{0.5}$ ($97\%$), competitive with strong VLA baselines despite going through an explicit STL specification rather than mapping observations directly to actions.

Table~\ref{tab:libero_pro_results} reports two perturbation types from LIBERO-PRO~\cite{zhou2025liberopro}: initial-position (Pos) perturbations, which reposition objects in the scene, and task (Task) perturbations, which swap the target object or goal while staying in-distribution.

Under LIBERO-PRO's perturbations (Table~\ref{tab:libero_pro_results}), every baseline collapses: OpenVLA and $\pi_0$ fall to $0\%$, $\pi_{0.5}$ drops to $0\%$--$38\%$, and the code-as-policy baseline \textsc{CaP-Agent0} reaches only $12\%$--$26\%$. \textsc{STeP} retains $54\%$--$89\%$ success across five of the six suite/perturbation combinations, well above the strongest baseline in each ($\leq38\%$ for position, $\leq18\%$ for task perturbations); the one exception is Goal-suite task perturbations, where it is roughly on par with \textsc{CaP-Agent0} ($16\%$ vs.\ $17\%$), likely because Goal-suite rephrasings vary the goal predicate itself (e.g., open vs.\ turn-on), the one axis a paraphrase can most easily flip. STeP's advantage on the other five combinations reflects a difference in mechanism, not just capacity: STeP's grasp pose comes from Contact-GraspNet applied to the current observation and its RRT/MPC path is planned fresh to the object's actual pose, so a shifted initial position does not by itself break execution; task-phrasing perturbations only have to survive a single VLM parsing step into predicates, which is comparatively robust to rewording. An end-to-end VLA, in contrast, has its action distribution conditioned on the pixel layout and instruction tokens seen during training, so shifting either off-distribution can silently degrade the policy with no signal of failure.

\subsection{RQ2: Precise Language Following}

\begin{figure}[t]
    \centering

    \includegraphics[width=0.9\linewidth]{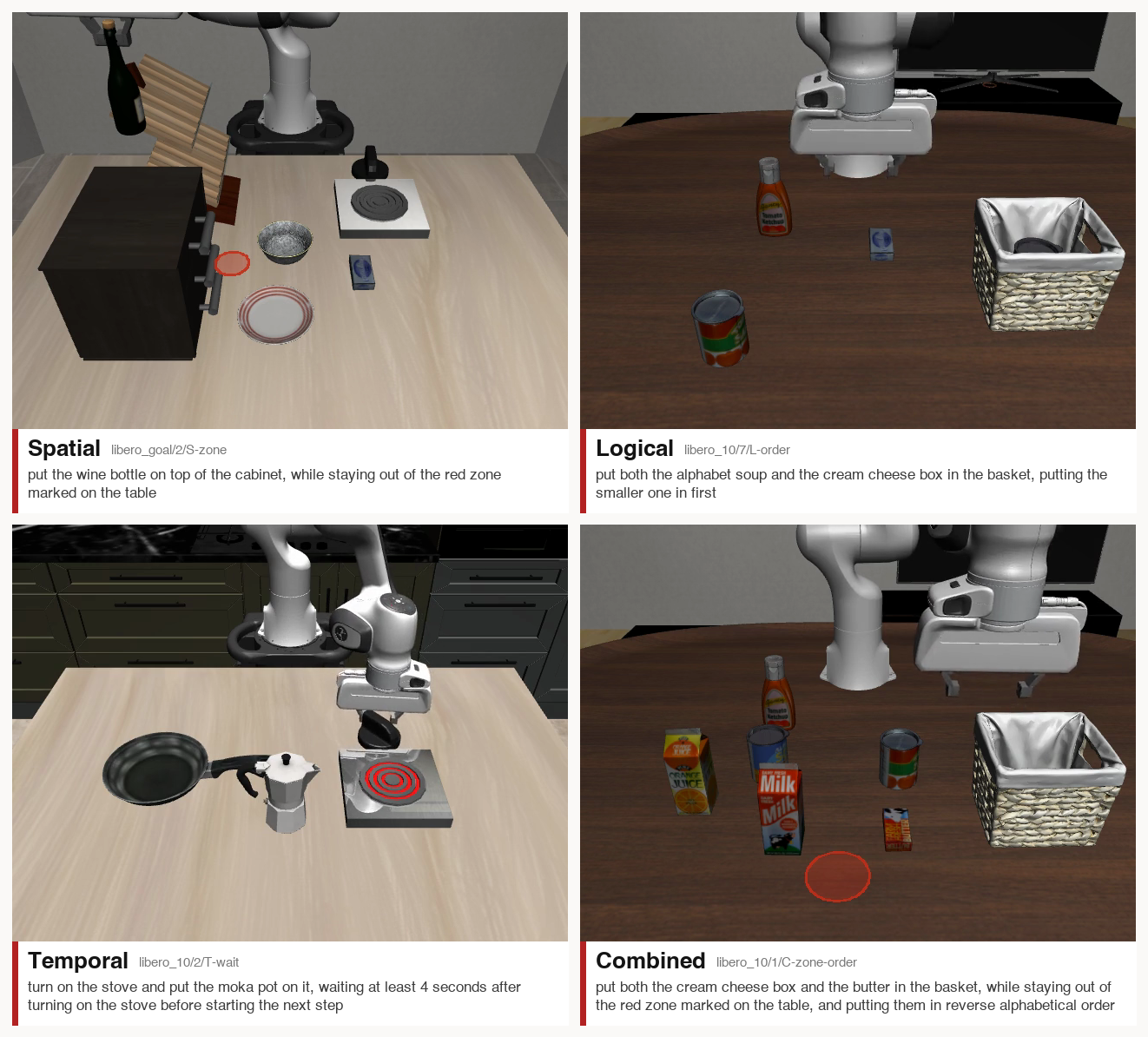}

    \vspace{0.4em}

    \resizebox{0.98\linewidth}{!}{%
    \begin{tabular}{l ccc ccc ccc}
    \toprule
    & \multicolumn{3}{c}{\textbf{SmolVLA}}
    & \multicolumn{3}{c}{\textbf{VLM-MPC-Cost}}
    & \multicolumn{3}{c}{\textbf{\textsc{STeP} (Ours)}} \\
    \cmidrule(lr){2-4}
    \cmidrule(lr){5-7}
    \cmidrule(lr){8-10}

    \textbf{Constraint}
    & \makecell{Task\\Succ. $\uparrow$}
    & \makecell{Viol.\\$\downarrow$}
    & \makecell{Safe\\Succ. $\uparrow$}
    & \makecell{Task\\Succ. $\uparrow$}
    & \makecell{Viol.\\$\downarrow$}
    & \makecell{Safe\\Succ. $\uparrow$}
    & \makecell{Task\\Succ. $\uparrow$}
    & \makecell{Viol.\\$\downarrow$}
    & \makecell{Safe\\Succ. $\uparrow$} \\
    \midrule

    Spatial
    & 0.48 & 0.48 & 0.22
    & 0.26 & 0.57 & 0.17
    & 0.87 & 0.26 & \textbf{0.74} \\

    Logical
    & 0.25 & 0.50 & 0.25
    & 0.67 & 0.22 & 0.67
    & 1.00 & 0.00 & \textbf{1.00} \\

    Temporal
    & 0.36 & 0.91 & 0.00
    & 0.08 & 0.83 & 0.00
    & 0.82 & 0.27 & \textbf{0.73} \\

    Combined
    & 0.00 & 0.14 & 0.00
    & 0.14 & 0.43 & 0.00
    & 0.86 & 0.14 & \textbf{0.86} \\

    \midrule
    \textbf{Overall}
    & \textbf{0.35} & \textbf{0.53} & \textbf{0.14}
    & \textbf{0.27} & \textbf{0.55} & \textbf{0.20}
    & \textbf{0.88} & \textbf{0.20} & \textbf{0.80} \\

    \bottomrule
    \end{tabular}%
    }

    \caption{\textsc{LIBERO-Constrained} benchmark and results.
    \textbf{Top:} representative spatial, logical, temporal, and combined
    constraint tasks. \textbf{Bottom:} comparison of SmolVLA,
    VLM-MPC-Cost, and \textsc{STeP}. Task success measures nominal task
    completion, violation rate measures constraint violations, and safe
    success requires both task completion and constraint satisfaction.}

    \label{fig:libero_constrained_results}
\end{figure}
\textbf{LIBERO-Constrained.}
Figure~\ref{fig:libero_constrained_results} (bottom) evaluates both nominal task
completion and satisfaction of the additional task requirements. We report
three metrics. \emph{Task Success} measures whether the underlying LIBERO
manipulation goal is completed, irrespective of constraint satisfaction.
\emph{Violation Rate} measures the fraction of executions in which at least
one specified constraint is violated. \emph{Safe Success} requires both
successful task completion and satisfaction of all specified constraints.
This distinction is important because an execution can complete the nominal
task while still violating the language instruction.

The main difference appears in the gap between task and safe success. Both
SmolVLA and VLM-MPC-Cost frequently violate the additional task requirements,
leading to substantially lower safe success, particularly for spatial,
temporal, and combined constraints. In \textsc{STeP}, these requirements are
represented explicitly in the active STL specification and remain available to
System~1 throughout execution, rather than being implicit in a learned policy
or encoded only as an MPC cost. Spatial requirements can therefore influence
geometric planning directly, while temporal and logical requirements remain
explicitly monitored throughout execution. Overall, \textsc{STeP} achieves
$80\%$ safe success, compared with $14\%$ for SmolVLA and $20\%$ for
VLM-MPC-Cost, while reducing the violation rate to $20\%$, from $53\%$ and
$55\%$, respectively.

\textbf{Real World.}
We additionally evaluate whether language-derived specifications transfer to
physical execution on a UR3e. The four task categories introduce spatial
exclusion regions, identity-based sorting, shared-workspace constraints, and
fragile-object handling. We compare \textsc{STeP} against
\textsc{VLM-MPC-Cost}, which retains the same high-level reasoning and
model-based execution structure but removes the explicit STL representation.
As shown in Figure~\ref{fig:real_world_combined}, \textsc{STeP} achieves
higher safe success across all four categories, with the largest improvement
on fragile-object handling.
\subsection{RQ3: Few-Shot Replanning}
\label{sec:rq3}

\begin{figure}[t]
    \centering
    \includegraphics[width=\linewidth]{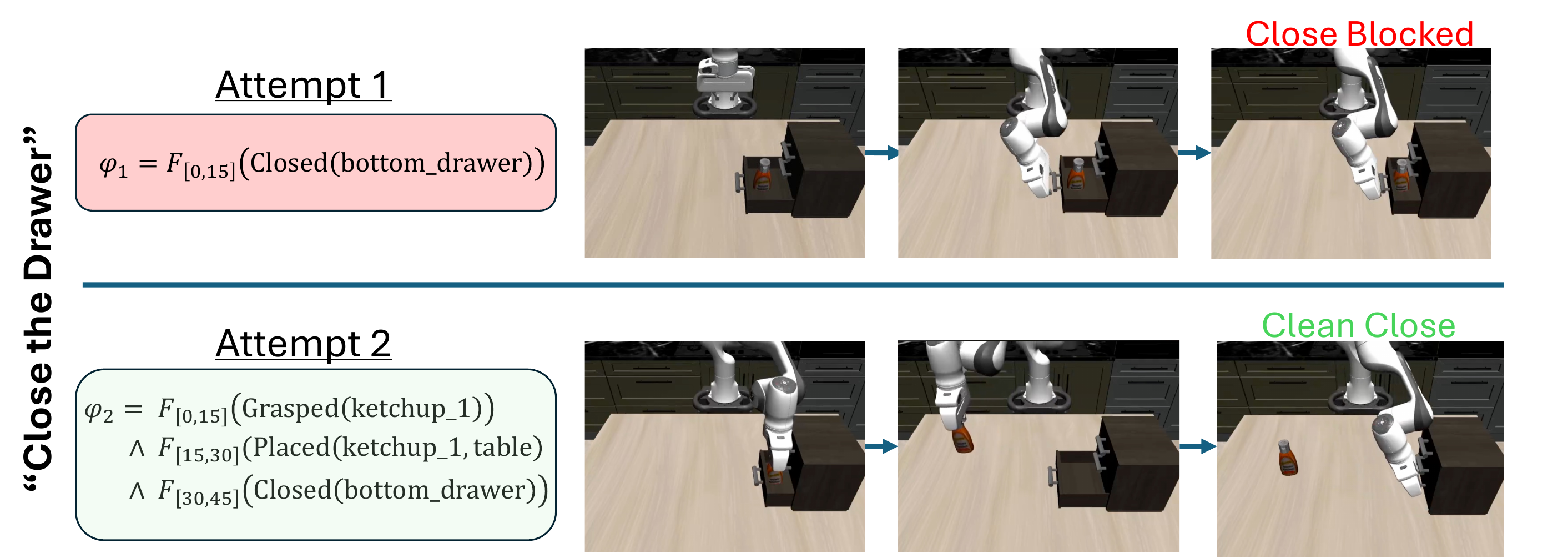}

    \vspace{5pt}
    \footnotesize
    \setlength{\tabcolsep}{4pt}
    \renewcommand{\arraystretch}{1.15}
    \begin{tabular}{lcccc}
    \toprule
    & \textbf{Succeeded} & \textbf{After 1} & \textbf{After 2} & \textbf{After 3} \\
    & \textbf{first try} & \textbf{recall} & \textbf{recalls} & \textbf{recalls} \\
    \midrule
    w/o Memory & 1/20 & 3/20 & 4/20 & 4/20 \\
    \textsc{STeP} (Ours) & 0/20 & 7/20 & 9/20 & 11/20 \\
    \bottomrule
    \end{tabular}
    \caption{Few-shot replanning experiment.
    \textbf{Top:} Example rollout showing how \textsc{STeP} replans after the
    initial drawer-closing attempt fails.
    \textbf{Bottom:} Cumulative success over 20 trials, each allowing up to three
    recalls. ``After $N$ recalls'' denotes the number of trials that succeeded
    using at most $N$ recalls, compared with an ablation without Memory
    (Recall Context).}
    \label{fig:recall_example}
\end{figure}

To evaluate few-shot replanning, we design a scenario intended to
induce failure of the direct closing plan: given the instruction ``close the
drawer,'' a ketchup bottle left inside the drawer physically blocks it from
closing (Figure~\ref{fig:recall_example}, left). System 1 first attempts the
direct \texttt{close} subtask, but the closing predicate's time window elapses
without being satisfied, triggering a recall to System 2. The improvement over
the no-Memory ablation comes from what that recall carries: the history log
retains the closing predicate's STL robustness trace over the failed attempt,
rather than only a binary success/failure signal. This allows System 2 to
observe that the drawer plateaued at a fixed, still-open position, which,
combined with the new observation, provides the context needed to infer that
the bottle must first be removed. Each trial is allowed up to three recalls
before being marked a failure; we run 20 trials and report cumulative success
as a function of the recall budget (Figure~\ref{fig:recall_example}, right).
With three recalls, \textsc{STeP} succeeds in 11/20 trials, compared with 4/20
without Memory, indicating that retaining execution context provides useful
information for correcting the failed plan.

\section{Conclusion}
\label{sec:conclusion}
We presented \textsc{STeP}, a specification-based agentic framework that
uses Signal Temporal Logic as a shared representation between language-based
task reasoning and low-level robot execution. Rather than treating language
instructions as informal prompts to a policy, \textsc{STeP} converts spatial,
temporal, and logical requirements into STL specifications that guide
model-based planning and control, are monitored during execution, and are
reused during replanning. System 1 primarily enforces these specifications
through model-based planning and control, while using a learned grasp-pose
network to propose candidate grasps. Across simulated manipulation and
real-world tabletop tasks, our results show that this formal interface
improves constraint satisfaction and enables more targeted recovery after
intermediate failures. These results indicate that language-conditioned
robots can benefit from an explicit specification layer between high-level
agentic reasoning and low-level control, rather than relying on a single
learned policy to implicitly satisfy all task requirements.
\section{Limitations}
\label{sec:limitations}

The current implementation handles temporal requirements most effectively at
the task-planning level. System~2 can explicitly schedule subtask windows,
whereas System~1 operates with a shorter receding horizon, so distant deadlines
have limited influence on the MPC objective until they enter that horizon.
This allows the controller to wait when ahead of schedule, but provides less
ability to recover once execution falls behind. \textsc{STeP} also assumes a fixed predicate library $\mathcal{P}$ and
predicate-to-skill map $\kappa$. This keeps specifications grounded in
reliably executable predicates and skills, but restricts the framework from generalizing to entirely new tasks. Future work could let System 2
propose new predicates on the spot, or learn new executors from experience.

\bibliographystyle{IEEEtran}
\bibliography{example}
\end{document}